\RequirePackage{fix-cm}
\documentclass{svjour3}                     
\smartqed  
\def\makeheadbox{}
\date{}
\usepackage{graphicx}
\usepackage{amsmath}
\usepackage{amsfonts}
\usepackage{booktabs}
\usepackage{makecell}
\usepackage{balance}
\usepackage[nolist]{acronym}
\usepackage{hyperref}
\usepackage{geometry}
\title{The FaceChannel: A Fast \& Furious Deep Neural Network for Facial Expression Recognition}

\author{Pablo~Barros$^{*}$, Nikhil~Churamani and~Alessandra~Sciutti}

\institute{Pablo Barros and Alessandra Sciutti \at
              Cognitive Architecture for Collaborative Technologies Unit, Istituto Italiano 
              di Tecnologia, Genova, Italy\\
              $^{*}$Correspondent author: \\
              \email{pablo.alvesdebarros@iit.it}
              \\ \and
              Nikhil Churamani \at 
              Department of Computer Science and Technology, University of Cambridge, United Kingdom 
              }
      
\begin{document}

\maketitle

\begin{abstract}

Current state-of-the-art models for automatic \acf{FER} are based on very deep neural networks that are effective but rather expensive to train. Given the dynamic conditions of \ac{FER}, this characteristic hinders such models of been used as a general affect recognition.  In this paper, we address this problem by formalizing the FaceChannel, a light-weight neural network that has much fewer parameters than common deep neural networks. We introduce an inhibitory layer that helps to shape the learning of facial features in the last layer of the network and thus improving performance while reducing the number of trainable parameters. To evaluate our model, we perform a series of experiments on different benchmark datasets and demonstrate how the FaceChannel achieves a comparable, if not better, performance to the current state-of-the-art in \ac{FER}. Our experiments include cross-dataset analysis, to estimate how our model behaves on different affective recognition conditions. We conclude our paper with an analysis of how FaceChannel learns and adapt the learned facial features towards the different datasets.

\keywords{Facial Expression Recognition \and Convolutional Neural Network \and Affective Computing}
\end{abstract}

\section{Introduction}

Evidence shows that humans can perceive, recognize, and commonly understand a set of `basic' emotions from facial expressions across cultures and around the world~\cite{Ekman1971}. Adapting an automatic \acf{FER} system to achieve such a capability, however, is still an open a difficult task. One of the most critical characteristics that yet needs to be addressed is how each person expresses the basic emotions differently, most of the time by combining different basic concepts or even shortly transitioning between them~\cite{cavallo2018emotion}. Understanding the compositionality of affect help us to understand better each other, and to derive a larger comprehension of affect than the ones posted by most of the current automatic \acf{FER} systems~\cite{Hamann2004,Hess2016}.

Given that most of the solutions for automatic \acf{FER} are difficult to adapt, due to their constrict computational nature, the most common solutions for this problem are to formalize affect into a way that bounds the categorization of such systems \cite{griffiths2003iii,barrett2006solving,Afzal2009}. Most of these systems, thus, are extremely restricted on what they can recognize as affect, given the availability of data to train them. But most importantly, the current state-of-the-art on \acf{FER} are deep neural networks with millions of parameters to tune. This implies they are extremely difficult to adapt to novel affect representation~\cite{mehta2018facial}.

Deep learning models usually learn how to represent affective features by updating filters based on a large number of data samples, using strongly supervised learning methods~\cite{hazarika2018self,huang2019speech,Kret2013,barros2018omg,kollias2019deep,kollias2020analysing}. As a result, these models can extract facial features for a collection of different individuals, which contributes to their generalization of expression representations enabling a universal and automatic \ac{FER} machine. The development of such models was supported by the collection of several ``in-the-wild'' datasets~\cite{dhall2012collecting,mollahosseini2017affectnet,zadeh2018multimodal,zafeiriou2017aff} that provided large amounts of well-labelled data. These datasets usually contain emotion expressions from various multimedia sources ranging from single frames to a few seconds of video material. Because of the availability of large amounts of training data, the performance of deep learning-based solutions forms the state-of-the-art in \ac{FER}, benchmarked on these datasets~\cite{choi2018convolutional,marinoiu20183d,du2019spatio,yang2018deep}.

Most of these models, however, employ large and deep neural networks that demand a high computational power for training and re-adapting \cite{krizhevsky2012imagenet,zheng2018multimodal,huang2019speech}. As a result, these models specialize in recognizing emotion expressions under conditions represented in the datasets they are trained with \cite{lindt2019facial,siqueira2020efficient}. Thus, when these models are used to recognize facial expression under different conditions, not represented in the training data, they tend to perform poorly. This is the case on several social applications, such as human-robot interaction \cite{tapus2019perceiving}. Retraining these networks to adapt to new application scenarios is usually the solution for adapting them. Yet, owing to the large and deep architecture of these models, retraining the entire network with changing conditions is rather expensive. 

Furthermore, once trained, these deep neural models are relatively fast to recognize facial expressions when provided with rich computational resources. With reduced processing power, however, such as in robotic platforms, these models usually are extremely slow and do not support real-time application.

\begin{figure*} 
	\center{\includegraphics[width=1\linewidth]{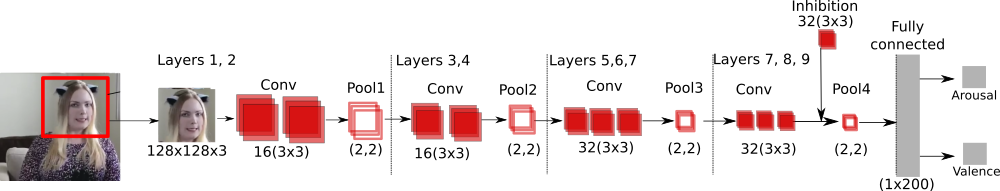}}
	\caption{The FaceChannel and all the parameters and details of the neural architecture.}
	\label{fig:faceChannel}
\end{figure*}

In this paper, we address the problem of very deep neural networks by formalizing the FaceChannel neural network. This model is an upgraded version of the Multi-Channel Convolution Neural Network, proposed in our previous work~\cite{barros2016developing}.

The FaceChannel is a light-weight convolution neural network, with around $2$ \textit{million} updatable parameters, that employs inhibitory layers trained from scratch. To evaluate our model, we perform a series of experiments on recognizing categorical and dimensional affect from different facial expressions datasets. We perform in-dataset and cross-dataset experiments, in order to evaluate in-depth the generalization of our model and its capability of adapting towards different scenarios. We also compare the performance of our model with current state-of-the-art deep neural networks for facial expression recognition, and demonstrate that the FaceChannel has a similar or better performance when compared to most of them, but with much fewer parameters to be updated. We also provide a discussion on how the FaceChannel consumes low computational resources to both train and adapt towards the different datasets. Further, we provide an analysis on how the features of the convolution layers of the FaceChannel are affected by each datasets' specific characteristics, and how they are shared on the adapting scenarios to boost the network's performance.

The FaceChannel was initially presented in our previous workshop paper \cite{Barrosface2020Ws}, and in this paper, we present an in-depth formalization of the model, extend the performance evaluations, including a novel set of cross-dataset assessments, and perform facial features learning analysis. Our goal with this paper is to complement our previous work with a detailed evaluation and understanding of the FaceChannel.

\section{The FaceChannel}

The FaceChannel presented here is an updated version of our previous work on facial expression recognition~\cite{barros2016developing}. We extend it by adapting the topology of the VGG16 model~\cite{VGG2016}, but with much fewer parameters to be trained, as exhibited in Fig.~\ref{fig:faceChannel}. Our model has a total of $10$ convolutional layers and $4$ pooling layers. Batch normalization is used after each convolutional layer and a dropout of with a $50\%$ chance is used after each pooling layer. Following our previous work, we adapted our last convolutional layer with shunting inhibitory connections~\cite{Fregnac2003} Each shunting neuron $S_{nc}^{xy}$ at the position ($x$,$y$) of the $n^{th}$ receptive field in the $c^{th}$ layer is activated as:

\begin{equation}
S_{nc}^{xy} = \frac{u_{nc}^{xy}}{a_{nc} + I_{nc}^{xy}}
\end{equation}

\noindent where $u_{nc}^{xy}$ is the activation of the convolution unit and $I_{nc}^{xy}$ is the activation of the inhibitory connections. A passive decay term, $a_{nc}$, which is learned, is shared among each shunting inhibitory connection. Each convolutional and inhibitory layer of the FaceChannel implements a \textit{ReLU} activation function.

Typically, in convolutional neural networks trained on images, the first layers learn how to represent and highlight edges, contours, and contrasts~\cite{yosinski2015understanding}. When trained to recognize faces, though, the last layers of such a network usually highlight facial characteristics, sometimes resembling even facial action units~\cite{mousavi2016understanding,zhou2017action}. The shunting neurons have a role of over-specifying the filters of the last layer. These neurons enhance the capability of the filters to extract strong and high-level facial representations, which improves the network capability of clustering database-specific features into the last layer~\cite{barros2016developing}. This improves the network generalization and makes it easier to be updated for a novel scenario, as and when needed.

The output of the convolutional layers is fed to a fully connected layer with $200$ units, each one implementing a \textit{ReLU} activation function, which is then fed to an output layer. Our model is trained using a \textit{categorical cross-entropy loss} function. 

As typical for most deep learning models, our FaceChannel has several hyper-parameters that need to be tuned. We optimized our model to maximize the recognition accuracy using a \ac{TPE}~\cite{bergstra2011algorithms} and use the optimal training parameters throughout all of our experiments. The entire network has around $2$ \textit{million} adaptable parameters, which makes it very light-weight as compared to commonly used VGG16-based networks.

\section{Experimental Setup}

To evaluate the FaceChannel, we perform several intra- and inter-dataset experiments. As some of these datasets do not contain enough data to train a deep neural network properly, we repeat three evaluation routines for each dataset:~(i)~we train the model using the indicated experimental protocol given by each dataset;~(ii)~we pre-train the model using the AffectNet dataset~\cite{mollahosseini2017affectnet} and evaluate it using the experimental protocol of each of the other datasets; and~(iii)~we pre-train the model using the AffectNet dataset, and fine-tune it using the training protocol for each of the evaluated datasets.

Our fine-tuning routine only trains the last fully connected layer of the FaceChannel. We performed empirical exploration experiments which demonstrated that re-training the entire network did not improve, and in some cases even decreased the networks' performance. Also, by fine-tuning only the last fully connected layer we guarantee that the facial features learned by the convolution layers are preserved, and only the decision-making on how to tune these features towards the labels of each dataset is affected.

Running these three experimental setups help us to investigate the capabilities of our model (a)~to learn facial features from each specific dataset;~(b)~to learn general features from a vast number of examples from the AffectNet dataset; and~(c)~ to adapt the learned features towards the individual specificity of each of the datasets.

\subsection{Datasets}

\begin{figure} 
	\center{\includegraphics[width=1\linewidth]{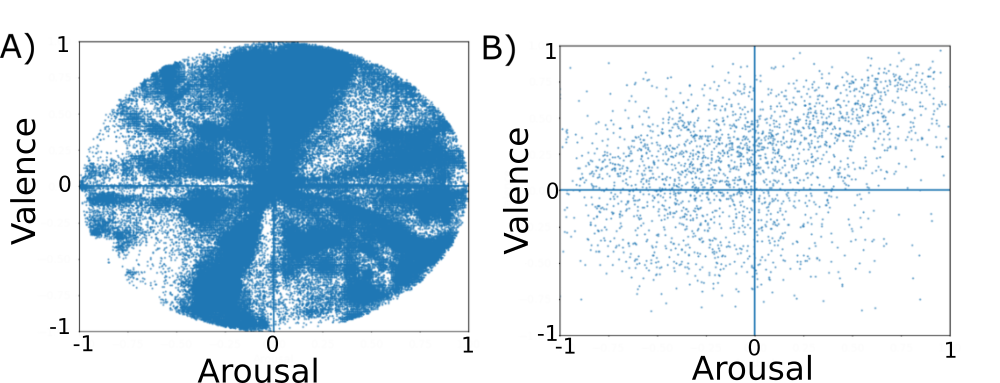}}
	\caption{Annotation distributions for: A) the AffectNet dataset~\cite{mollahosseini2017affectnet} has a high variance on arousal and valence with a large number of data points and B) The continuous expressions of the OMG-Emotion~\cite{barros2018omg} videos cover high arousal and valence spread.}
	\label{fig:dataDistribution}
	\end{figure}

\subsubsection{AffectNet}	

The AffectNet~\cite{mollahosseini2017affectnet} dataset consists of more than $1$ \textit{million} images obtained from web crawllers. Approximately half of them were manually annotated and contain a single label based on a continuous arousal and valence. All our experiments involving this dataset are performed using the training and validation subset separation, as the test-set labels are not publicly available. 

The AffectNet has the most representative data distribution among all the datasets we experimented with (as illustrated in Fig. ~\ref{fig:dataDistribution} A). Given the amount of samples and the data distribution, we use it to pre-train the FaceChannel in two of our experimental routines. This guarantees that the FaceChannel has enough data samples to learn general facial features.

\subsubsection{OMG-Emotion}

The \acf{OMG-Emotion} dataset~\cite{barros2018omg} is composed of $675$ videos with around $10$ hours of data, with an average length of one minute per video of persons performing monologues. The videos were gathered using web crawlers and manually annotated based on a continuous arousal and valence scale. Evaluating the FaceChannel on this dataset helps us to assess how our model performs when recognizing affect from particular individuals in a continuous scenario. Each video displays one person, and it is annotated at utterance level. The emotion expressions displayed in the \ac{OMG-Emotion} dataset are heavily impacted by person-specific characteristics that are highlighted by the gradual change of emotional behavior over the entire video. The videos in this dataset cover a diverse range of arousal and valence values as seen in Fig.~\ref{fig:dataDistribution} B).

\subsubsection{FER$+$}

We also evaluate the FaceChannel on the FER$+$ dataset~\cite{BarsoumICMI2016}. The FER+ contains around $31,000$ face images crawled from the internet. Although collected in a similar way as the AffecNet dataset, the labels of the FER$+$ were collected using a different strategy. Each image was annotated by $10$ different annotators using a categorical selection based on the \textit{basic emotions} (Angry, Disgust, Fear, Happy, Sad, Surprise) and Neutral. The categorical distribution for all $10$ annotators is used to create a single label per image. Each label, thus, represents a composition of the basic emotions. It is important for us to evaluate FaceChannel on this dataset to assess its capability of adapting to composed labels.

\subsubsection{FABO}

To evaluate the FaceChannel in a controlled environment setting, which is not present in any of the previous datasets, we use the Bi-modal Face and Body benchmark dataset FABO~\cite{Gunes2006}. This corpus is composed of images with the face and body posture of different subjects. In our experiments, we focus on the facial expressions only. The dataset is composed of different videos, and in each of them, one subject performs a pre-defined expression in a cycle of two to four expressions per video.

\begin{table}[h]
    \center\begin{tabular}{ c c | c c}
        \toprule
        Emotional State & Videos & Emotional State & Videos      \\ \midrule
        Anger           & $60$  & Happiness       & $28$  \\ 
        Anxiety         & $21$  & Puzzlement      & $46$  \\ 
        Boredom         & $28$  & Sadness         & $16$  \\ 
        Disgust         & $27$  & Surprise        & $13$  \\ 
        Fear            & $22$  & Uncertainty     & $23$  \\\bottomrule
    \end{tabular} 
    \caption{Number of videos available for each emotional state in the FABO dataset. Each video has $2$ to $4$ executions of the same expression.}

\label{tab_numberOfFrames}
\end{table}
 
 Each of the videos of the FABO dataset has an annotation of the apex of the expressions. Six individual observers annotated each video and a voting process was executed. A total number of $281$ videos are used as elaborated in Table \ref{tab_numberOfFrames}. Ten affective labels were used: ``\textit{Anger}", ``\textit{Anxiety}", ``\textit{Boredom}", ``\textit{Disgust}", ``\textit{Fear}", ``\textit{Happiness}", ``\textit{Surprise}", ``\textit{Puzzlement}", ``\textit{Sadness}" and ``\textit{Uncertainty}". We only used the apex frames for each of the expressions to train our model.
 
\subsection{Training and Evaluation Protocol}

We follow the training and test separation established by the authors each of the datasets in our experimental setup. We enforce this, as it is important to maintain the comparability with previously proposed models. For each frame, we run a face detection based on the DLib python~\footnote{https://pypi.org/project/dlib/} library and re-size it to a dimension of $128$x$128$ pixels.

\subsection{Metrics}

We use two different metrics to measure the performance of the FaceChannel in our experiments: \textit{accuracy}, when recognizing emotion labels, and the \acf{CCC}~\cite{Lawrence1989} between the outputs of the models and the true labels when recognizing arousal and valence. The \ac{CCC} is computed as:

\begin{equation}
CCC = \frac{2 \rho \sigma_x \sigma_y}{\sigma_{x}^2 + \sigma_{y}^2 + (\mu_x - \mu_y)^2}
\label{eq:ccc}
\end{equation}

\noindent where $\mu_x$ and $\mu_y$ represent the mean for model predictions and the annotations and $\sigma_{x}^2$ and $\sigma_{y}^2$, are the corresponding variances. $\rho$ is Pearson's Correlation Coefficient between model prediction labels and the annotations.

The \ac{CCC} allows us to compare directly the annotations available in the AffectNet and \ac{OMG-Emotion} datasets. While the \textit{accuracy} helps us to evaluate the performance of the FaceChannel on the FABO and FER+ datasets. 

\section{Results}

\subsection{AffectNet}
Although, the AffectNet corpus is very popular, not many researchers report the performance of arousal and valence prediction on its validation set. This is probably the case as most of the research uses the AffectNet dataset to pre-train neural models for generalization tasks in other datasets, without reporting the performance on the AffectNet itself.  We report the final results on the AffectNet dataset in Table~\ref{tab:affectnet}. The baseline provided by the authors uses an AlexNet-based Convolutional Neural Network~\cite{krizhevsky2012imagenet} re-trained to recognize arousal and valence. A similar approach is reported by Hewitt and Gunes~\cite{hewitt2018cnn}, but using a much reduced neural network, to be deployed on a smart-device. Lindt et al. \cite{lindt2019facial} report experiments using the VGGFace, a variant of the VGG16 network pre-trained for face identification. Kollias et al. \cite{kollias2020deep} proposed a novel training mechanics, where it augmented the training set of the AffectNet using a Generative Adversarial Network (GAN), and obtained the best reported accuracy on this corpus, achieving $0.54$ CCC for arousal and $0.62$ CCC for valence.  Our FaceChannel provides an improved performance when compared to most of these results achieving a \ac{CCC} of $0.46$ for arousal and $0.61$ for valence. Different from the work of Kollias et al. \cite{kollias2020deep}, we train our model using only the available training set portion, and expect these results to improve when training on an augmented training set.

\begin{table}[h]
    \center 
    \begin{tabular}{ c |  c | c  }\toprule
    
    \textbf{Model}                          & \textbf{Arousal}  & \textbf{Valence} \\\toprule
        
        AlexNet~\cite{krizhevsky2012imagenet}   & 0.34              & 0.60   \\
        MobileNet~\cite{hewitt2018cnn} & 0.48 & 0.57  \\
        VGGFace~\cite{lindt2019facial}  & 0.40 & 0.48 \\
        \textbf{VGGFace+GAN~\cite{kollias2020deep}}  & \textbf{0.54} & \textbf{0.62} \\
        
        Face Channel                   & 0.46     & 0.61 \\\midrule
        
    \end{tabular} 
    \caption{\acf{CCC}, for arousal and valence when evaluating the FaceChannel with the AffectNet dataset.}

\label{tab:affectnet}
\end{table}

\subsection{OMG-Emotion}
Training the FaceChannel on different datasets makes it easier to visualize the impact of a large number of training samples, as reported in Table \ref{tab:OMG-Facechannel}. When trained only with the OMG-Emotion dataset, the model achieved a CCC of $0.12$ for arousal and $0.23$ for valence, which is much lower than any other training configuration. Training the model with the AffectNet dataset increased drastically its performance, but still pre-training the model with the AffectNet and fine-tuning it with the OMG-Emotion train set to yield the highest performance. This demonstrates that although the features learned when training the model with the AffectNet are general and somehow more reliable than the ones learned with the OMG-Emotion alone, the specificities of the OMG-Emotion data are still beneficial to improve the models' performance. 

 The performance of the FaceChannel is very similar when compared to the current state-of-the-art results on the \ac{OMG-Emotion} dataset, as reported by the winners of the OMG-Emotion challenge where the dataset was proposed~\cite{zheng2018multimodal,peng2018deep,deng2018multimodal} as exhibited in Table~\ref{tab:OMG}. All these models also reported the use of pre-training of uni-sensory convolutional channels to achieve such results, but employed deep networks with much more parameters to be fine-tuned in an end-to-end manner. The use of attention mechanisms~\cite{zheng2018multimodal} to process the continuous expressions in the videos presented the best results of the challenge, achieving a \ac{CCC} of $0.35$ for arousal and $0.49$ for valence. Temporal pooling, implemented as bi-directional \acfp{LSTM} , achieved the second place, with a \ac{CCC} of $0.24$ for arousal and $0.43$ for valence. The late-fusion of facial expressions, speech signals, and text information reached the third-best result, with a \ac{CCC} of $0.27$ for arousal and $0.35$ for valence. The complex attention-based network proposed by Huang~et~al.~\cite{huang2019speech} was able to achieve a \ac{CCC} of $0.31$ in arousal and $0.45$ in valence, using only visual information.

\begin{table}[h]
    \center 
    \begin{tabular}{ c |  c | c | c | c  }\toprule
    
         \textbf{Model} &  \textbf{Trained} & \textbf{Tuned}                        & \textbf{Arousal}  & \textbf{Valence} \\\toprule
          FaceChannel &OMG-Emotion & -                             &  0.12             & 0.23 \\
        FaceChannel &AffectNet & -                             &  0.25             & 0.31 \\
         \textbf{FaceChannel} &AffectNet & OMG-Emotion                             &  0.32             & 0.46 \\\hline
        
    \end{tabular} 
    \caption{\acf{CCC}, for arousal and valence when evaluating the different versions of the trained FaceChannel with the OMG-Emotion dataset.}

\label{tab:OMG-Facechannel}
\end{table}
\begin{table}[h]
    \center 

    \begin{tabular}{ c |  c | c  }\toprule
    
         \textbf{Model}                          & \textbf{Arousal}  & \textbf{Valence} \\\toprule
        \textbf{Zheng~et~al.~\cite{zheng2018multimodal}} & \textbf{0.35} & \textbf{0.49} \\ 
        
        Huang~et~al.~\cite{huang2019speech}     & 0.31              & 0.45 \\
        Peng~et~al.~\cite{peng2018deep}         & 0.24              & 0.43 \\    
        
        Deng~et~al.~\cite{deng2018multimodal}   & 0.27              & 0.35 \\ 
        FaceChannel                             &  0.32             & 0.46 \\
          \hline
        
    \end{tabular} 
    \caption{\acf{CCC}, for arousal and valence when evaluating the best version of the FaceChannel with the OMG-Emotion dataset.}

\label{tab:OMG}

\end{table}

\subsection{FER+}
Similar to what happened when training the model with the OMG-Emotion, the routine that includes pre-training the FaceChannel with the AffectNet and fine-tuning it with a specific dataset achieved the best results when evaluating the FER+ dataset, as reported in Table \ref{tab:FER-all}. In this case, however, the performance gain is not as underlined as it was in the OMG-Emotion dataset, probably as the FER+ dataset has already a large number of data samples for training. The most important difference here is on the label representation, as the FER+ represents the labels using a distribution of annotations over the entire category range. This is probably the most impacting change when fine-tuning the FaceChannel with the FER+ causes, which impacts directly on the performance of the model.

 \begin{table}[h]
    \center 

    \begin{tabular}{ c |  c | c | c  }\toprule
    
         \textbf{Model} &\textbf{Trained} &   \textbf{Tuned}                      & \textbf{Accuracy}\\\toprule
         Face Channel & FER+ &- & 87.50\%  \\
  Face Channel & AffectNet & - & 88.20\%  \\
        \textbf{Face Channel} & AffectNet & FER+ & \textbf{90.50\%}  \\\midrule  
    \end{tabular} 
    \caption{Accuracy when evaluating the different versions of the trained FaceChannel with the FER+ dataset.}

\label{tab:FER-all}

\end{table}
 
\begin{table}[h]
    \center 
    \begin{tabular}{ c |  c  }\toprule
    
         \textbf{Model}                          & \textbf{Accuracy}\\\toprule
        CNN VGG13~\cite{BarsoumICMI2016}  & 84.98\%  \\
        SHCNN~\cite{miao2019recognizing} & 86.54 \\
        TFE-JL~\cite{li2018facial}& 84.3\\
        ESR-9~\cite{siqueira2020efficient}&87.15 \\
    
        \textbf{Face Channel} & \textbf{90.50\%}  \\\midrule  
    \end{tabular} 
    \caption{Accuracy when evaluating the FaceChannel with the FER+ dataset.}

\label{tab:FER}

\end{table}
When trained and evaluated with the FER+ model, our FaceChannel provides improved results as compared to those reported by the dataset authors~\cite{BarsoumICMI2016}. They employ a deep neural network based on the VGG13 model, trained using different label-averaging schemes. Their best results are achieved using the labels as a probability distribution, which is the same strategy we used. We outperform their result by almost $6\%$ as reported in Table \ref{tab:FER}. We also outperform the results reported in Miao et al.~\cite{miao2019recognizing}, Li et al.~\cite{li2018facial}, and Siqueira et al.~\cite{siqueira2020efficient} which employ different type of complex neural networks to learn facial expressions.

\subsection{FABO}
Table \ref{tab:FABO-all} reports the experiments of the different versions of the FaceChannel on the FABO dataset. Similar to the experiments involving the FER+ dataset, the results of training the Face-Channel only with the FABO dataset are not much worst than when training with only the AffectNet dataset. The FABO dataset also contains enough data to tune the convolution layers towards learning facial features. The improvement obtained when training the model with the AffectNet and tuning it with the FABO dataset, however, demonstrates that the FABO dataset also has specific peculiarities which are not depicted by the FaceChannel when trained with the AffectNet alone. As in all of our previous experiments, the best results were obtained in this training configuration.
\begin{table}[h]
    \center 
    \begin{tabular}{ c |  c | c | c }\toprule
    
         \textbf{Model} &  \textbf{Trained} &\textbf{Tuned}                          & \textbf{Accuracy}\\\toprule
         Face channel & FABO & - &76.2\% \\
         Face channel & AffectNet & - &75.9\% \\
        \textbf{Face channel} & AffectNet & FABO & \textbf{80.54\%} \\\bottomrule
        
    \end{tabular} 
    \caption{Accuracy when evaluating the different versions of the trained FaceChannel with the FABO dataset.}

    \label{tab:FABO-all}
\end{table}
Our model achieves higher accuracy for the experiments with the FABO dataset as well when compared with the state-of-the-art for the dataset~\cite{Chen2013}. They report an approach based on recognizing each video-frame, similar to ours. The results reported by Gunes~et~al.~\cite{Gunes2009} for Adaboost and SVM-based implementations are reported using a frame-based accuracy. Our FaceChannel outperforms both models, as illustrated in Table \ref{tab:FABO}.

  \begin{table}[h]
    \center 
    \begin{tabular}{ c |  c  }\toprule
    
         \textbf{Model}                          & \textbf{Accuracy}\\\toprule
        \makecell{Temporal Normalization~\cite{Chen2013}}           &  66.50\%     \\ 
        
        Bag~of~Words~\cite{Chen2013}           & 59.00\%   \\
        SVM~\cite{Gunes2009}~           & 32.49\%   \\    
        
         Adaboost~\cite{Gunes2009}           & 35.22\%   \\ 
        \textbf{Face channel} & \textbf{80.54\%} \\\bottomrule
        
    \end{tabular} 
    \caption{Accuracy when evaluating the FaceChannel with the FABO dataset.}

\label{tab:FABO}
\end{table}

To better understand how the FaceChannel classifies the emotions on the FABO dataset, we present as well the accuracy per classification class in Table \ref{tab:FABOClass}. We observe that the network provides an stable classification over all the classes, including the ones with less examples such as the ``Surprise" class. 

  \begin{table}[h]
    \center 
    \begin{tabular}{ c |  c  }\toprule
    
         \textbf{Class}                          & \textbf{Accuracy}\\\toprule
        Anger &	75.8\\
        Anxiety&	77.8\\
        Boredom&	80.1\\
        Disgust&	78.3\\
        Fear&	85.1\\
        Happiness&	83.9\\
        Puzzlement&	85.4\\
        Sadness&	80.4\\
        Surprise&	75.8\\
        Uncertainty &	77.4 \\\bottomrule
    \end{tabular} 
    \caption{Class specific accuracy when evaluating the FaceChannel with the FABO dataset.}

\label{tab:FABOClass}
\end{table}

\section{Discussions}
 
Our experiments demonstrate how the FaceChannel can perform better than most deeper neural networks for recognizing facial expressions on the AffectNet, OMG-Emotion, FER+ and FABO datasets. Besides reaching a good performance when evaluating facial expressions alone, the advantage of our model is its smaller configuration when compared to the popular versions of convolution neural networks.
 
\subsection{Fast training} 
The final configuration of the FaceChannel has around $2$ \textit{million} parameters to be updated during training. The entire model was developed using Keras~\cite{gulli2017deep}, and trained on a system with an Intel i7 CPU, $16$GB of RAM, and a Quadro RTX $4000$ GPU with $8$GB of memory. During pre-training, all the $2$ \textit{million} parameters were updated, while during fine-tuning only the last fully-connected layer was re-trained, reducing the number of trainable parameters to 800 thousand.

\begin{table}[h]
    \center 

    \begin{tabular}{ c | c | c | c | c  }\toprule

        \textbf{Dataset} & \textbf{N. Samples} & \textbf{Parameters}  & \textbf{GPU} & \textbf{Training time} \\ \toprule
         AffectNet & 1 million & 2 millions & yes & 6 hours \\
         AffectNet & 1 million & 2 millions & no & 10 hours \\
         
         FER+ & 97 thousand & 2 millions & yes & 2 hours \\
         FER+ & 97 thousand & 2 millions & no & 6 hours \\
         
         FABO & 5 thousand & 2 millions & yes & 45 minutes \\
         FABO & 5 thousand & 2 millions & no & 3 hours \\
        
         OMG-Emotion & 10 thousand & 2 millions & yes & 1 hour \\
         OMG-Emotion & 10 thousand & 2 millions & no & 5 hours \\\hline
        
    \end{tabular} 
    \caption{Training time, number of parameters and number of training samples for all of our experiments when training the FaceChannel for 100 epoches.}

\label{tab:training-time}

\end{table}
Table \ref{tab:training-time} reports the training time for all of our full-training experiments. We observe that, as expected, the training times when using the GPU are much smaller than when using the CPU. But the most important to note is that even when training with the $1$ \textit{million} examples of the AffectNet dataset, the FaceChannel took only $10$ hours to train with the CPU, and $6$ hours to train with the GPU. When compared to most of the state-of-the-art deep learning models~\cite{li2020deep}, these numbers demonstrate a great benefit of a smaller and light-weighted deep neural architecture.

The training effort is even smaller when we compare the resources needed for fine-tuning the network, reported in Table \ref{tab:tuning-time}. With fewer parameters to be updated, the network takes only $2$ hours to train on the $97$ thousand examples of the FER+ dataset. Also, the fine-tuning achieved the highest results in each of these datasets, demonstrating the capability of the FaceChannel to adapt robustly, but also quickly, towards new representations present on affective label re-associations.
 
  \begin{table}[h]
    \center 
    \begin{tabular}{ c | c | c | c | c  }\toprule

        \textbf{Dataset} & \textbf{N. Samples} & \textbf{Parameters}  & \textbf{GPU} & \textbf{Training time} \\ \toprule

         FER+ & 97 thousand & 800 thousand & yes & 45 minutes \\
         FER+ & 97 thousand & 800 thousand & no & 2 hours \\
         
         FABO & 5 thousand & 800 thousand & yes & 10 minutes \\
         FABO & 5 thousand & 800 thousand & no & 40 minutes \\
        
         OMG-Emotion & 10 thousand & 800 thousand  & yes & 25 hours \\
         OMG-Emotion & 10 thousand & 800 thousand  & no & 1 hour \\\hline

    \end{tabular} 
    \caption{Training time, number of parameters and number of training samples for all of our experiments when fine-tuning the FaceChannel for 100 epoches.}

\label{tab:tuning-time}
\end{table}

\begin{table}[h]
    \center 
    \begin{tabular}{ c | c  }\toprule

        \textbf{Model} & \textbf{Parameters} \\ \toprule

         FaceChannel & 2 Million \\
                  MobileNet & 4.9 Million \\
         VGG13+ & 34 Million \\
                  AlexNet & 60 Million \\
         VGGFace+ & 138 Million\\\hline
        
    \end{tabular} 
    \caption{Number of trainable parameters for deep learning models discussed in our results section.}

\label{tab:ParameterNumbers}
\end{table}

We also provide in Table \ref{tab:ParameterNumbers}, a comparison of how many parameters the deep learning models presented on on our experiments have. We observe that the FaceChannel has by far the lowest number of parameters although presenting a better or same performance as the other models, as exhibited in Section 4.

\begin{figure} 
	\center{\includegraphics[width=1\linewidth]{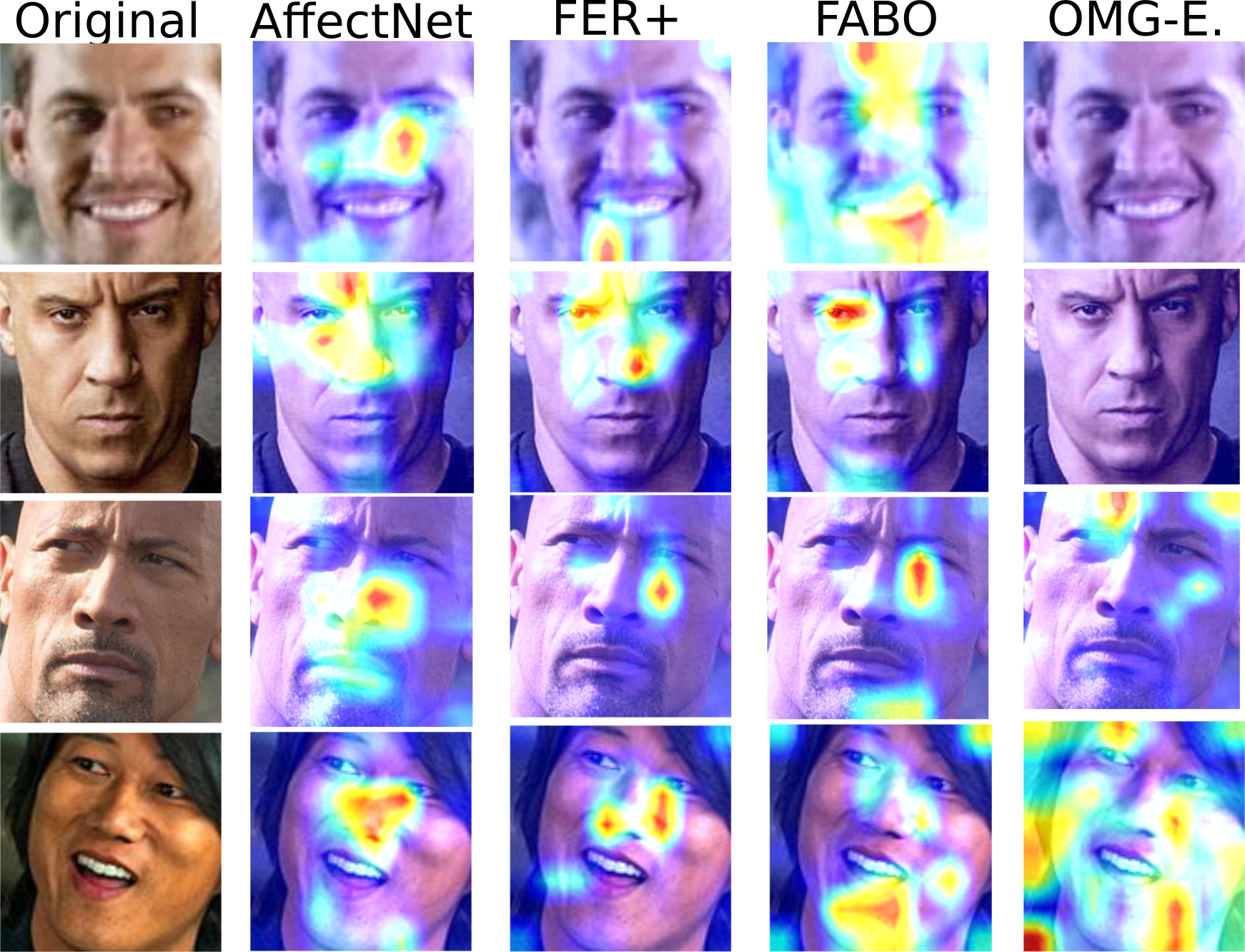}}
	\caption{Feature-level analysis of how the different datasets impact the FaceChannel's capability of detecting facial features. The figure illustrates the results of the GradCam \cite{selvaraju2017grad} visualization technique on the FaceChannel trained with the different datasets.}
	\label{fig:faceFeatures}
	\end{figure}
\subsection{Facial features adaptation} 

To better understand the impact that each of these datasets has on the convolution filters of the FaceChannel we perform a visualization analysis, based on the GradCam method~\cite{selvaraju2017grad}. Figure~\ref{fig:faceFeatures} exhibits the neural activation of the last convolutional layer of the FaceChannel when trained with each of the datasets. We observe that the network trained with the AffectNet and FER+ datasets focuses mostly on features on the central area of the face, mostly encapsulating the eyes/nose/mouth region. This is mostly due to the images in these datasets having a centralized pre-processing. The networks trained with the FABO dataset, however, have a strong bias towards focusing on the eyes and the chin. This is mostly due to the training samples on this dataset which are composed of extremely exaggerated facial expressions, in particular on the eyes/chin areas. The network trained with OMG-Emotion dataset, however, does not present a unique pattern, but a much more spread neural activation. Combined with the performance results, we can affirm that this happens because the network was not able to learn any specific feature characteristics when trained with this dataset. 

Besides explaining the performance of the network, these analyses also help us understand why pre-training with the AffectNet achieves the best results on all of these datasets. The combination of the number of training samples, with a large variety of ``in-the-wild" expressions for sure helped the network to tune its filters towards general facial features. This effect is observed also by recent research on training deep neural networks with facial expressions~\cite{mousavi2016understanding,zhou2017action,patel2020facial}.

That our network was able to learn general features, even with a light-weighted architecture, is another testament to its strength at quick adaptation towards novel scenarios.

\section{Conclusions}

We presented, in this paper, the formalization of the FaceChannel for automatic \acf{FER}. Our neural network has an architecture based on the VGG16, but optimized to use much fewer trainable parameters. The reduction of the computational costs implies on a faster adaptation towards new scenarios, which is common when recognizing affect from different persons.

Our experiments demonstrate that our neural network has a compatible, and in most cases even better, performance than the current state-of-the-art models for automatic \acf{FER}. We also demonstrate, using different \ac{FER} datasets with specific data characteristics, how our model can be quickly adapted and fine-tuned for specific affective perception scenarios. To guarantee the reproducibility and dissemination of our model, we have made it fully available on GitHub\footnote{\url{https://github.com/pablovin/FaceChannel}}.

We plan to study and deploy our model on platforms with reduced data processing capabilities, such as social robots. Also, we believe that the light-weighted architecture will allow a quick adaptation towards individual aspects of affective performance, and thus, the development of personalized solutions is encouraged.

\section{Compliance with Ethical Standards}

\noindent\textbf{Funding} This project is supported by a Starting Grant from the European Research Council (ERC) under the European Union’s Horizon 2020 research and innovation programme. G.A. No 804388, wHiSPER".

\noindent\textbf{Conflicts of interest/Competing interests.} The authors declare that they have no conflict of interest.

\noindent\textbf{Availability of data and material.} All the datasets used in our experiments are publicly available.

\noindent\textbf{Code availability.} The entire code for the network's topology and the trained network are available at: \url{https://github.com/pablovin/FaceChannel}

\bibliographystyle{ieeetr}
\bibliography{main.bib}

\begin{acronym}
\acro{CCC}{Concordance Correlation Coefficient}
\acro{FER}{Facial Expression Recognition}
\acro{LSTM}{Long Short-Term Memory}
\acro{OMG-Emotion}{One Minute Gradual Emotion Recognition}
\acro{TPE}{Tree-structured Parzen Estimator}
\end{acronym}
\end{document}